\documentclass[%
 reprint,
 amsmath,amssymb,
 aps,
]{revtex4-2}

\usepackage{graphicx}
\usepackage{dcolumn}
\usepackage{bm}
\usepackage{mathptmx}

\begin{document}

\preprint{APS/123-QED}

\title{Measuring Trustworthiness or Automating Physiognomy? A Comment on Safra, Chevallier, Gr{\`e}zes, and Baumard (2020)}
\thanks{Correspondence regarding this article should be addressed to Rory Spanton, School of Psychology, University of Plymouth, Drake Circus, UK, PL4 8AA. E-mail: rory.spanton@plymouth.ac.uk \\
R.S. drafted the manuscript and created Figure 1. R.S. and O.G. conceived the content of the manuscript, and edited the manuscript. The authors declare no competing interests. \\
We thank Iris van Rooij, Sam Forbes and Brad Wyble for helpful comments on a previous draft of this manuscript.}

\author{Rory W. Spanton}
\affiliation{
 School of Psychology, University of Plymouth, \\ United Kingdom}

\author{Olivia Guest}
\affiliation{Donders Institute for Brain, Cognition and Behaviour, Radboud University, \\ Netherlands}



\date{\today}


\maketitle


%

Interpersonal trust --- a shared display of confidence and vulnerability toward other individuals --- can be seen as instrumental in the development of human societies. Safra, Chevallier, Gr{\`e}zes, and Baumard (2020) \cite{Safraetal2020} studied the historical progression of interpersonal trust by training a machine learning (ML) algorithm to generate trustworthiness ratings of historical portraits, based on facial features. They reported that trustworthiness ratings of portraits dated between 1500--2000CE increased with time, claiming that this evidenced a broader increase in interpersonal trust coinciding with several metrics of societal progress. We argue that these claims are confounded by several methodological and analytical issues and highlight troubling parallels between Safra et al.'s algorithm and the pseudoscience of physiognomy. We discuss the implications and potential real-world consequences of these issues in further detail.

\section*{Defining and Measuring Trustworthiness}

Although Safra, Chevallier, Gr{\`e}zes, and Baumard (2020) \cite{Safraetal2020} (henceforth Safra et al.) make strong claims about displays of trustworthiness in humans, they do not clearly define this construct. Crucially, the authors conflate perceived trustworthiness (an observer's subjective evaluation of trust toward a face) and trusting behaviour. These constructs are demonstrably separate. While one may display more trustworthy facial expressions to appear perceivably trustworthy, this does not logically dictate that a person who is perceived as trustworthy must want to behave in a trustworthy manner \cite{vanRooij2020Blog}. Facial expressions can map on to many cognitive states; somebody can smile while in discomfort, or because they are genuinely pleased to see you. People can also be deceptive, and physical displays of trustworthiness do not necessarily indicate genuinely trusting or cooperative behaviour \cite{DePaulo1996Lying}. 

However, the authors conclude that their results are ``suggestive of an actual shift in social trust'' between 1500--2000CE. This implies that their results have implications for human behaviour, rather than for individual perceptions of trustworthiness. The authors have since sought to clarify that their algorithm exclusively rates perceived trustworthiness in a retroactive disclaimer in their OSF repository. However, the ambiguity in their article remains, obfuscating their aims and implying their algorithm measures something it cannot. Indeed, many features of perceived trustworthiness emerge as a highly dynamic product of a given society during a given historical period, and not simply from facial features. Even on the level of the individual, displays of trust are affected by many other factors such as personality and life experience \cite{Delhey2003}. Yet by analysing facial features alone, Safra et al.'s algorithm is detached from these factors, greatly undermining their conclusion that social trust increased over time.

Safra et al. also present limited statistical evidence for their key claims that trustworthiness increases alongside various metrics of societal development. For instance, they write in their abstract that trustworthiness displays parallel the decline in interpersonal violence and an uptake in democratic values between 1500--2000CE. However, they present no primary statistical analyses to support the former association. They also report that the latter association between predicted trustworthiness and democratic values was not significant in one of their test sets of portraits, and was not robust to the addition of time as a covariate in the other. These results cannot be taken as support for the claims made in their abstract.

\begin{figure*}
    \centering
    \includegraphics[width=\textwidth]{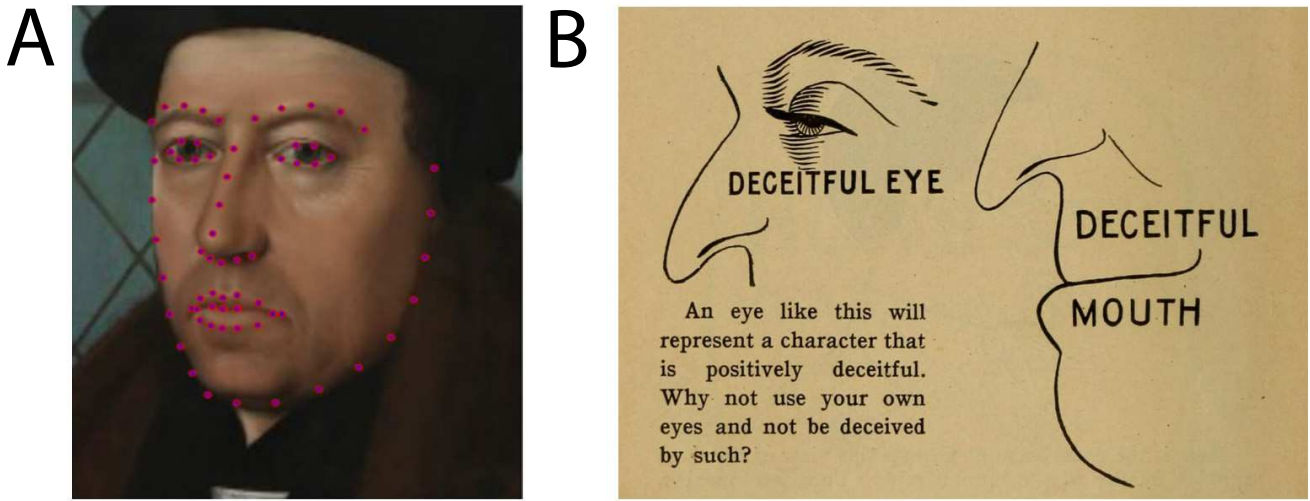}
    \caption{A visual comparison of A) a portrait judged to have low trustworthiness by Safra et al.'s algorithm, and B) illustrations from an early 20th Century physiognomy guide \cite{Vaught1902}. Image A originally from Safra et al. (2020), reproduced under CC BY 4.0 rights; Image B reproduced under public domain rights.}
    \label{fig:label}
\end{figure*}

\section*{Physiognomy}

The authors' reliance on facial features to measure trustworthiness also leads to another, arguably even greater issue. By inferring human behaviour from facial features, Safra et al.'s trustworthiness algorithm constitutes a direct example of physiognomy.

Physiognomy is the pseudoscientific theory that personality and behaviour can be predicted by the physical characteristics of the human body \cite{Porter2003Science}; most notably facial features. This idea was first formalised by Greek writers in the 5\textsuperscript{th} Century BCE, but surged in popularity during the 18\textsuperscript{th} and 19\textsuperscript{th} Centuries CE \cite{Graham1961PhysPopular}. Physiognomy was also a core part of Nazi ideology and helped fuel their ethnic persecution and eugenics during the Second World War \cite{Gray2004PhysNazism}. Prompted by growing scientific scepticism of phrenology and the post-war collapse of Nazism, critics labelled physiognomy a pseudoscience \cite{Porter2003Science}.

Despite rejection from the scientific community, physiognomy is experiencing a rebirth in contemporary ML research \cite{Birhane2020Decolonising}. Although immediate character judgements based on a subject's face are often erroneous \cite{Zebrowitz2017FirstImpsRev}, recent ML algorithms have nonetheless attempted to use facial characteristics to make judgements about criminality \cite{Wu2017PhysCriminality}, attractiveness \cite{Wu2016PhysAttractiveness}, and sexual orientation \cite{Leuner2019PhysSexualOrientation}. Despite their more modern approaches, these articles share the broad goals and assumptions of historical physiognomy. Yet, these historical parallels are rarely acknowledged \cite{Birhane2020Decolonising}.

This is also true of Safra et al.'s algorithm. At its core, the algorithm identifies facial action units and computes a trustworthiness rating based upon their physical properties. This trustworthiness rating then informs conclusions about the attitudes and behaviours of people throughout European history. The algorithm therefore produces an automated physiognomic judgement (see Figure 1), transforming an assessment of facial characteristics into a prediction of behavioural characteristics and future social outcomes \cite{Stark2021Physiognomic}. Even if the algorithm only reflects perceptions of trustworthiness, it still systematizes this human bias by judging facial features, as does physiognomy. Given physiognomy's history as a pseudoscientific vehicle for racism and eugenics, this is problematic for the scientific and ethical integrity of Safra et al. (2020).

\section*{Algorithmic Bias}

Going further than a historical parallel, Safra et al's algorithm could potentially produce judgements that reflect racial prejudices, due to algorithmic bias. No ML algorithm is a neutral transformer of data to output, and many types of bias contribute to its predictions \cite{Danks2017}. Though not acknowledged, such bias is present in many stages of the authors' model development.

Firstly, Safra et al. trained their algorithm to imitate the judgements of human participants. This choice was intentional, motivated by the authors' aim to generate human-like trustworthiness evaluations. However, the algorithm's ratings of real faces have a weak correlation with human trustworthiness judgements (\textit{r} = .22), suggesting it struggles to generalise to real stimuli as opposed to maximally distinctive training faces \cite{Steed2021Predictive}. This means it is unlikely that the algorithm's behaviour reflects human judgements as intended. Furthermore, most of the human participants whose behaviour the algorithm mirrored were Western, Educated, Industrialised, Rich, and Democratic (WEIRD; \cite{Henrich2010WEIRD}). Although this is typical in psychological research, it is problematic here because the authors explicitly generalise their algorithm's judgements to historical populations with large demographic and cultural differences. In reality, it is unlikely that many people in historical Europe would have shared the trustworthiness evaluations of a modern WEIRD sample.

Secondly, the authors trained their algorithm using a set of exemplar face avatars that were exclusively white, meaning all the features the algorithm learned to associate with trustworthiness are those common in white faces. Although no algorithm is free from bias, this particular choice could make the algorithm show a bias against faces of marginalised ethnic groups, taking it even closer to an automated form of physiognomy.

One might argue that because the authors analysed predominantly white portrait sitters in their key tests, training their algorithm on only white faces was a principled decision. However, this overlooks other issues. The first is that this selection of test portraits constitutes another form of bias in itself. In most broad historical populations, only a very select few individuals would have had the financial means or social status to have their portraits painted and preserved for generations. Due to systemic social inequalities throughout history, these individuals were predominantly white men who likely had more homogeneous facial characteristics than a broader multicultural sample from a given time period. Despite this, Safra et al. go on to infer psychological attributes of a wider population based on their evaluation of a deliberately pro-male, wealthy, white sample. As such, these inferences are likely invalid.

As a result of unchecked bias, the authors' algorithm could make harmful judgements about minoritised ethnic individuals that have lasting and damaging consequences, if applied in a non-scientific context. This is especially troubling when considering that there are similar ML algorithms currently being used to automate high-stakes social judgements. For instance, algorithms with biases against people of colour are used to predict criminal activity, social welfare eligibility, and even for ethnic persecution \cite{Birhane2019Algorithmic}. As Safra et al.'s algorithm makes judgements in a similar domain, a generalized form of their algorithm could conceivably be used for similar purposes by a malicious actor. This constitutes an ethical risk that, along with comparable research, jeopardises the trustworthiness of ML itself, underlining how this field struggles to educate its researchers to avoid the mistakes of the past.

\section*{Conclusion}

In sum, Safra et al.'s conclusion that trustworthiness increased over time alongside metrics of societal development is not supported by strong evidence and is based upon flawed assumptions and methods. Moreover, this research raises serious ethical issues as a modern example of physiognomy and could ultimately lead to harmful consequences for marginalised groups and individuals. To mitigate these potential harms, it is crucial that researchers, reviewers, publishers, and funding bodies are critical of physiognomic principles and question strong claims based on weak statistical evidence when developing and evaluating ML research. Failure to do so would reflect poorly on the entire discipline.



\bibliography{apssamp}

\end{document}